\pdfoutput=1
\PassOptionsToPackage{table,xcdraw}{xcolor}
\documentclass[11pt]{article}

\usepackage[preprint]{acl}

\usepackage{times}
\usepackage{latexsym}

\usepackage[T1]{fontenc}
\usepackage[utf8]{inputenc}

\usepackage{microtype}

\usepackage{inconsolata}

\usepackage{graphicx}

\usepackage[table,xcdraw]{xcolor}
\usepackage{multirow}
\usepackage{comment}
\usepackage{booktabs}
\usepackage{lipsum}  

\newcommand{\todo}[1]{\textcolor{red}{TODO: #1}}

\usepackage{array}

\usepackage{tcolorbox}
\newcounter{promptno}[section]
\newlength\mystoreparindent
\newenvironment{prompt}[1][]
{
  \setlength{\mystoreparindent}{\the\parindent}
  \setlength{\parindent}{0pt}
  \refstepcounter{promptno}
  \par\medskip
  \noindent
  \begin{tcolorbox}[left=1pt,right=1pt]
  \textsc{{template \small\thesubsection.\thepromptno}}\\
  \small
  \tt
}{
  \end{tcolorbox}
  \setlength{\parindent}{\mystoreparindent}
  \medskip
}

\title{From Templates to Natural Language: Generalization Challenges in Instruction-Tuned LLMs for Spatial Reasoning}

\author{%
Chalamalasetti Kranti${^\mathbf{1}}$, Sherzod Hakimov${^\mathbf{1}}$, David Schlangen${^\mathbf{1,2}}$\\$^{\mathbf{1}}$Computational Linguistics, Department of Linguistics\\
University of Potsdam, Germany\\
$^{\mathbf{2}}$German Research Center for Artificial Intelligence (DFKI), Berlin, Germany\\
{\texttt{\{kranti.chalamalasetti, sherzod.hakimov, david.schlangen\}@uni-potsdam.de}}
}

\begin{document}
\maketitle
\begin{abstract}
Instruction-tuned large language models (LLMs) have shown strong performance on a variety of tasks; however, generalizing from synthetic to human-authored instructions in grounded environments remains a challenge for them. In this work, we study generalization challenges in spatial grounding tasks where models interpret and translate instructions for building object arrangements on a $2.5$D grid. We fine-tune LLMs using only synthetic instructions and evaluate their performance on a benchmark dataset containing both synthetic and human-authored instructions. Our results reveal that while models generalize well on simple tasks, their performance degrades significantly on more complex tasks. We present a detailed error analysis of the gaps in instruction generalization.

\end{abstract}

\section{Introduction}
Accurate spatial grounding~\citep{DBLP:conf/iros/GreenEWL06, DBLP:conf/ro-man/Brenner07} is important for effective human-robot interaction~\citep{bisk-etal-2016-natural, DBLP:conf/rss/ShridharH18, DBLP:conf/icra/HatoriKKTTUKT18}. It involves the robot interpreting spatial references~\citep{DBLP:conf/iros/HuttenrauchEGT06} and relational cues expressed in natural language instructions~\citep{DBLP:conf/robot/Dang-VuPR15, DBLP:journals/ijrr/PaulAARH18, DBLP:journals/arcras/TellexGKM20, DBLP:conf/emnlp/BiskHTABCLLMNPT20}. Large Language Models (LLMs) are increasingly used in robotics pipelines~\citep{DBLP:conf/corl/IchterBCFHHHIIJ22, DBLP:conf/iclr/WangLYSBQWX024, DBLP:conf/case/LimPEPLK24, macaluso2024toward} to translate such instructions into high-level plans and then into low-level robot actions. While these models have shown progress in instruction-following tasks, they continue to face challenges~\citep{DBLP:journals/tacl/0001EC23, DBLP:conf/iros/Ding0P023, DBLP:journals/corr/abs-2505-10872} when grounding spatial language references.

Human instructions are often abstract~\citep{minsky1980k}, underspecified~\citep{hendriks2014referential}, and context-dependent. For example, a user might say \textit{``put the red block on top of that blue one''} in one case, and \textit{``make a staircase''} in another. These variations rely on shared context and conceptual inference, making them harder to interpret~\citep{ch-etal-2024-retrieval, chaturvedi-etal-2024-nebula} than synthetic or templated instructions, which are designed to be direct and explicit.

\begin{figure}[t]
\centering
  \includegraphics[width=0.38\textwidth]{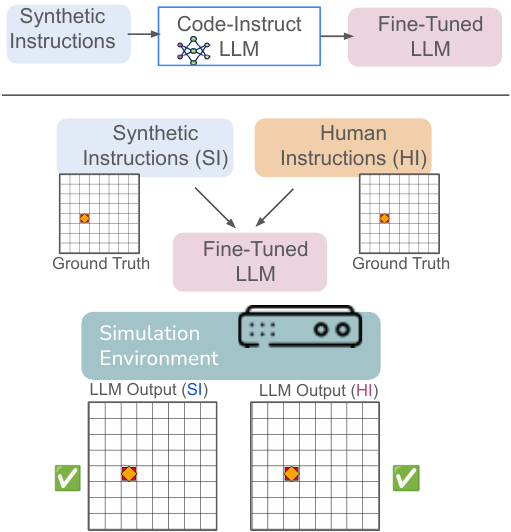}
  \caption{Fine-tuned LLMs trained on synthetic instructions are evaluated on both synthetic (SI) and human (HI) instructions. A virtual simulator verifies whether generated response matches the gold configuration.
  }
  \label{fig:firstpagefigure}
\end{figure}

To explore these issues in a controlled setting, we investigate them in the context of a structure-building task, where instructions guide the manipulation of objects on a grid. We leverage an existing dataset~\citep{DBLP:journals/corr/abs-2409-11041} that includes both synthetic and human-authored instructions for the same grounded spatial reasoning task. Each task involves building specific arrangements of colored shapes on an 2.5D grid with size $8\times8$, referred to as \textit{``boards.''} Each board is paired with both synthetic and human-authored instructions, enabling direct comparison of model behavior (see Figure~\ref{fig:firstpagefigure}) across instruction styles within a shared task space.

We evaluate how well pre-trained LLMs perform  with both synthetic and human-authored instructions across different board types. Furthermore, we fine-tune instruction-tuned LLMs using only the synthetic instructions and evaluate them whether they generalize to human-authored ones. Our results show that while fine-tuning improves model performance, and models achieve high execution scores on simple boards, performance gains are limited on regular boards. This suggests that instruction generalization is closely tied to the compositional and structural complexity of the task, and that fine-tuning solely on synthetic data is insufficient for robust transfer to instructions that require more abstract or relational understanding.

Our contributions are as follows: (a) we evaluate pre-trained and fine-tuned LLMs on both synthetic and human-authored instructions, and report quantitative performance across different board types (simple vs. regular); (b) we conduct a detailed qualitative analysis of model responses before and after fine-tuning, categorizing the types of errors across instruction styles and board complexity; (c) we analyze the embedding similarity of human-authored instructions, to their synthetic counterparts, the number of shapes involved, and how these factors correlate with model performance; and (d) we test the generalization of fine-tuned models on two related, out-of-domain code generation tasks, and report the performance related to the generalization.

\section{Related Work}

\paragraph{Spatial Grounding with LLMs} Spatial grounding refers to the task of interpreting spatial references in natural language and linking them to a target environment. Recent work has explored the use of LLMs for tasks such as block manipulation~\citep{DBLP:journals/corr/abs-2302-06706, DBLP:journals/corr/abs-2409-17126}, table setting~\citep{DBLP:conf/icra/LiangHXXHIFZ23, DBLP:journals/access/VempralaBBK24}, tool usage~\citep{xu2023creative} and collaborative assembly~\citep{DBLP:conf/case/LimPEPLK24, macaluso2024toward, joglekar2024towards}. Our work is inspired by these efforts but differs in how the grounding is performed. Prior approaches generate sequences of atomic actions or low-level code that are directly executable in the environment. In contrast, we focus on generating abstract functions that encode spatial semantics but are not immediately grounded; these functions are interpreted and executed by a downstream system. This abstraction allows us to analyze how well LLMs encode spatial structure independently of immediate action execution.

\paragraph{Instruction-to-Code Translation for Spatial Tasks} Instruction-to-code translation focuses on converting natural language inputs into executable programs or structured representations~\citep{DBLP:conf/icra/SinghBMGXTFTG23, DBLP:conf/icra/HuangMZB23, DBLP:conf/corl/HuangWZL0023, DBLP:journals/tmlr/WangX0MXZFA24, hu2024robo}. Our work extends these efforts by comparing model performance across two instruction styles, synthetic and human-authored, for the same spatial task. This setup enables us to study generalization and instruction-following behavior under consistent task conditions.

\paragraph{Generalization from Synthetic to Human Instruction} Instruction-tuned models often struggle to generalize well from synthetic to natural language~\citep{DBLP:conf/emnlp/LiZL023, nwankwo2025reli, DBLP:journals/corr/abs-2502-19417, DBLP:journals/corr/abs-2503-14023}. This reflects the challenges of handling variation in human-authored instructions. We extend this to a spatial grounding task and report similar failures in generalization despite high performance on synthetic tasks.

\paragraph{Fine-tuning on Synthetic vs. Human Instructions} Prior work has explored fine-tuning LLMs for collaborative structure building in environments like \textsc{Minecraft}~\citep{ch-etal-2024-retrieval, chaturvedi-etal-2024-nebula}. These studies typically fine-tune on a mix of synthetic and human-authored instructions and report limited generalization. Furthermore, \textsc{Minecraft} dataset~\citep{narayan-chen-etal-2019-collaborative} does not distinguish based on complexity and structures with and without object repetitions, making it difficult to isolate the impact of fine-tuning. Our work builds on these by analyzing performance gaps across instruction styles and board types that affect generalization.

\begin{figure}[t]
\centering
  \includegraphics[width=\columnwidth]{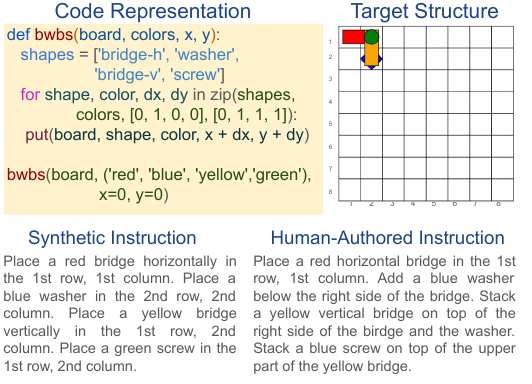}
  \caption{A simple board example showing how code (left) maps to a target structure (right), with both synthetic and human-authored instructions describing the same configuration. Models are expected to generate both the function definition and its usage based on the given instruction.}
  \label{fig:instruction_type_example_sb}
\end{figure}

\section {Task Formulation and Datasets}
\begin{figure}[t]
\centering
  \includegraphics[width=\columnwidth]{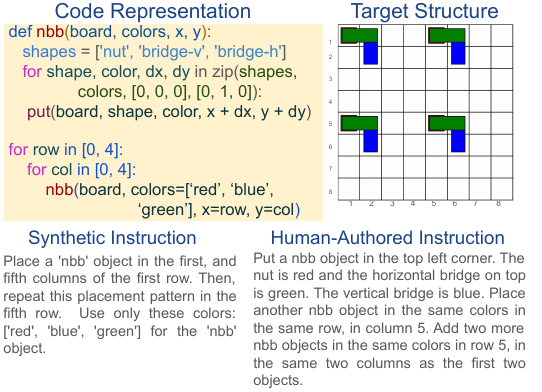}
  \caption{A regular board example illustrating repeated structural patterns defined by code (left), their corresponding target structure (right), and matching synthetic and human-authored instructions. The function definition is provided in the prompt; models are expected to generate only the usage code (e.g., nested for loops) based on the instruction.}
  \label{fig:instruction_type_example_rb}
\end{figure}

\textbf{Task}: it is framed as a two-player game between a programmer and a cobot (collaborative robot),
where there is no intermediate feedback and execution happens only at the end of the interaction. The programmer (instruction giver) is given a target board and instructs the cobot (instruction follower) to translate the instruction into code. The code generated by the cobot is executed in a virtual simulation environment and evaluated for correctness.

\textbf{Boards}: We use the SARTCo~\citep{DBLP:journals/corr/abs-2409-11041} dataset for finetuning and the main testbed for our experiments. It consists of $260$ human-authored instructions for a structure building task on a $2.5$D~\footnote{A $2$D grid with stacking support in each cell and without the complexity of full $3$D simulation.} grid. Each instance in the dataset includes a triplet: \textit{a code representation}, \textit{a target board}, and \textit{a natural language instruction}. An example consisting of the target board, an abstract Python function, and two styles of instructions, is shown in Figure~\ref{fig:instruction_type_example_sb} and Figure~\ref{fig:instruction_type_example_rb}. Each target board is a specific arrangement of objects on a grid. An \textit{object} is a composition of shapes such as \textit{washers}, \textit{screws}, \textit{nuts}, and \textit{bridges}. Boards are labeled as either \textit{simple}, with non-repetitive arrangements, or \textit{regular}, with repetitive arrangements. The goal is to generate this semantically grounded function from the instruction. This setup reflects instruction-following scenarios and enables evaluation of how well LLMs can spatially ground such instructions.

\textbf{Instructions}: they are of two types: synthetic \& Human-authored. Synthetic instructions are generated using a template-based grammar. They are structurally consistent and semantically explicit. They are available in both single-turn and multi-turn formats. Human-authored instructions are free-form language inputs for the same target structure. They are available only in single-turn format. Both types are aligned to the same structure building task.

\textbf{Learning on Synthetic Instructions}: we prompt LLMs with both instruction types and evaluate the performance by executing the generated code in a virtual simulation. The output is compared to the target to compute execution success, defined as the proportion of cases where the model output matches the target board. We fine-tune models on synthetic instructions and evaluate them on both synthetic and human-authored instructions. This setup tests how well models generalize from synthetic to human language in the same task domain.

\textbf{Cross-domain Datasets}: We also evaluate cross-domain generalization using the \textsc{HEXAGONS}~\citep{lachmy-etal-2022-draw} and \textsc{TidyBot}~\citep{DBLP:conf/iros/WuAKLZSBRF23} datasets. These datasets help assessing whether fine-tuned models are able to perform on different but related spatial instruction tasks.

\section{Experimental Setup}
We evaluate a diverse set of LLMs with varying sizes and architectures, focusing on their ability to interpret grounded spatial instructions and generate executable action sequences. These models are used as-is (few-shot) and fine-tuned on synthetic data only, without exposure to human-authored instructions during training. We use the \textit{clembench}~\citep{chalamalasetti-etal-2023-clembench} framework to manage player interactions.

\subsection{Evaluated Models \& Parameters}
\textbf{Models}: We include both code-centric models (\textsc{Qwen2.5-Coder}-7B, 32B) and general-purpose, instruction-following models (\textsc{Qwen3}-8B, 32B; \textsc{LLaMA-3.1}-8B, 3.3-70B). All models are loaded in 4-bit quantized precision using the \textsc{Unsloth} library\footnote{\url{https://unsloth.ai/}}. We compared these models against the commercial ones: \textsc{GPT-4o} and \textsc{Claude-4-Sonnet}.

\textbf{Fine-Tuning Details}: Models are fine-tuned on synthetic instruction–code pairs, where the input is a synthetic natural language instruction and the output is the corresponding Python code representation of the board. The training set~\citep{DBLP:journals/corr/abs-2409-11041} comprises a mixture of  simple boards ($1072$ boards) and regular boards ($1168$ boards). We adopt a chat-style prompt format, including context and environment details (more details in Appendix~\ref{sec:appendix-prompt-templates}), followed by the instruction and target code. We use the \textsc{QLoRA} configuration provided by \textsc{Unsloth} with the following hyperparameters: \textit{$3$ epochs}, \textit{$20$ steps}, \texttt{adam-8bit} optimizer, rank $r=16$, 
\texttt{lora\_alpha}=$16$, \texttt{lora\_dropout}=$0.10$, batch size=$8$, and learning rate=$1\mathrm{e}{-4}$.

\textbf{Hyperparameter Selection and Ablations}: To determine optimal fine-tuning settings, we conducted a comprehensive ablation study varying data composition, ordering, and hyperparameters. We evaluated models trained on (i) only simple boards, (ii) only regular boards, and (iii) combined boards with/without shuffling. Additionally, we explored the impact of training sample sizes ($100$-to-$1000$ per board type), batch sizes ($2$-to-$8$), learning rates ($1\mathrm{e}{-4}$, $2\mathrm{e}{-4}$), epoch counts ($1$-to-$5$), \texttt{lora\_dropout} ($0.10$, $0.15$, $0.20$) and early stopping patience($2$, $3$). These studies informed our final configuration: combined dataset with shuffled samples, full training set, batch size 8, 3 epochs, learning rate $1\mathrm{e}{-4}$, and dropout $0.10$. More details and ablation study results are available in Appendix~\ref{subsubsec:appendix-finetuning-ablationstudy}.

\subsection{Evaluation Metrics}
We assess model performance on instruction-following capability, execution accuracy, and cross-domain robustness.

\paragraph{Error Rate:} A metric capturing instruction adherence based on predefined response constraints (output formatting) in the prompt (Appendix~\ref{sec:appendix-prompt-templates}). In the clembench setup, responses that violate these constraints are marked as \texttt{abort}.

\paragraph{Execution Success Rate:} Our primary task success metric. Each model-generated function is executed in a virtual simulator and compared against the target board (an $8\times8$ grid with shape type, color, position, and stacking order). A prediction is successful if the resulting board configuration exactly matches the ground truth in component type, color, spatial placement, and stacking sequence.

\paragraph{Human Baseline:}
To contextualize model performance, we include a human-generated baseline. We developed a simple user interface (see Figure~\ref{fig:human_reconstruct_interface} in Appendix) that shows the human-authored instructions (by a different annotator) on the left and the setup to reconstruct the target board on the right. 
We hired an annotator and asked to reconstruct target boards. These outputs were executed using the same virtual simulator and computed the same \textit{Execution Success Rate} metric, allowing direct comparison between model and human performance.

\paragraph{Cross-domain robustness:} We measure generalization by comparing model performance before and after fine-tuning. Specifically, we evaluate whether models trained on synthetic instructions generalize to (1) human-authored instructions in the same task domain (structure building: synthetic $\rightarrow$ human), and (2) structurally similar but domain-shifted tasks such as drawing hexagons (\textsc{HEXAGONS}) and sorting objects (\textsc{TidyBot}).

\begin{comment}
\begin{table}
  \centering
  \footnotesize
  \begin{tabular}{ccccc}
    \hline
    \multirow{2}{*}{\textbf{Model}} & \multicolumn{2}{c}{\textbf{Simple Boards}} & \multicolumn{2}{c}{\textbf{Regular Boards}} \\
    & Before & After & Before & After \\
    \hline
    \texttt{Qwen2.5-Coder-7B}     & $0.07$ & $0.40$ & $-$ & $0.35$\\
    \texttt{Llama3.1-8B}     & $-$ & $0.08$ & $-$ & $0.21$\\
    \texttt{Qwen3-8B}     & $0.00$ & $0.09$ & $-$ & $0.12$\\
    \texttt{Qwen2.5-Coder-32B}     & $0.14$ & $\mathbf{0.68}$ & $-$ & $\mathbf{0.54}$\\    
    \texttt{Qwen3-32B}     & $0.00$ & $0.53$ & $-$ & $0.38$\\    
    \texttt{Llama3.3-70B}     & $0.11$ & $\mathbf{0.69}$& $-$ & $0.33$\\    
    \hline
    \texttt{GPT-4o}     & $0.72$ & $-$ & $0.55$ & $-$\\
    \texttt{Claude-4}     & $0.88$ & $-$& $0.33$ & $-$\\    
    \hline
    \texttt{Human-Baseline}     & $0.98$ & - & $0.76$ & -\\        
    \hline    
  \end{tabular}
  \caption{Performance of closed models (GPT-4o and Claude-4) along with the human baseline on simple boards (SB) and regular boards (RB).\todo{update with latest results}}
  \label{tab:results_ha}
\end{table}
\end{comment}

\begin{table*}
  \centering
  \begin{tabular}{lcccc|cccc}
    \hline
    \multirow{3}{*}{\textbf{Model}} &\multicolumn{4}{c}{\textbf{Abort Rate $\downarrow$}} & \multicolumn{4}{c}{\textbf{Performance $\uparrow$}} \\
    & \multicolumn{2}{c}{\textbf{SB}} & \multicolumn{2}{c}{\textbf{RB}} & \multicolumn{2}{c}{\textbf{SB}} & \multicolumn{2}{c}{\textbf{RB}} \\
    & Before & After & Before & After & Before & After  & Before & After  \\
    \hline
    \texttt{Qwen2.5-Coder-7B}     & $0.00$ & $0.00$ & $1.00$ & $0.00$ & $0.07$ & $0.40$ & $0.07$ & $0.35$ \\
    \texttt{Llama3.1-8B}     & $0.00$ & $0.11$ & $0.00$ & $0.09$ & $0.00$ & $0.08$ & $0.00$ & $0.21$ \\
    \texttt{Qwen3-8B}          & $1.00$ & $0.38$ & $0.95$ & $0.59$ & $0.00$ & $0.09$ & $0.00$ & $0.12$\\
    \texttt{Qwen2.5-Coder-32B}  & $0.00$ & $0.00$  & $0.00$ & $0.00$ & $0.14$ & $\mathbf{0.68}$ & $0.23$ & $\mathbf{0.54}$\\    
    \texttt{Qwen3-32B}          & $0.95$ & $0.00$ & $0.97$ & $0.00$ & $0.00$ & $0.53$ & $0.00$ & $0.38$\\
    \texttt{Llama3.3-70B}     & $0.00$ & $0.00$ & $0.72$& $0.06$ & $0.11$ & $\mathbf{0.69}$ & $0.07$ & $0.33$ \\    
    \hline
    \texttt{GPT-4o}     & $0.00$ & $-$ & $0.00$ & $-$ & $0.72$ & $-$ & $0.55$ & $-$\\
    \texttt{Claude-4(sonnet)}     & $0.00$ & $-$& $0.00$ & $-$ & $0.88$ & $-$ & $0.60$ & $-$\\    
    \hline
    % \texttt{Human-Baseline}     & $-$ & $-$ & $-$ & $-$ &   $0.98$ & $-$ & $0.76$ & $-$\\       
    \texttt{Human-Baseline}     & $-$ & $-$ & $-$ & $-$ &   \multicolumn{2}{c}{\textbf{0.98}} & \multicolumn{2}{c}{\textbf{0.76}}\\       
    \hline    
  \end{tabular}
  \caption{Abort rate and task performance (accuracy) on simple boards (SB) and regular boards (RB) from human-authored set, before and after fine-tuning on the synthetic set. While some models show gains on SB, improvements on RB are low.}
  \label{tab:results_ha}
\end{table*}

\section{Results}\label{sec:results}
We present quantitative and qualitative results, followed by cross-domain generalization analysis. Quantitative results are organized first by board type (simple and regular) for human-authored instructions, and then by instruction type (synthetic and human-authored) for overall comparison.

\begin{table}
  \centering
  \footnotesize
  \begin{tabular}{lcccc}
    \hline
    \multirow{2}{*}{\textbf{Model}} & \multicolumn{2}{c}{\textbf{FSG}} & \multicolumn{2}{c}{\textbf{FSC}} \\
    & SB & RB & SB & RB \\
    \hline
    \texttt{Qwen2.5-Coder-7B}     & $0.25$ & $0.25$ & $0.23$ & $0.19$\\
    \texttt{Llama3.1-8B}     & $0.02$ & $0.28$ & $0.11$ & $0.29$\\
    \texttt{Qwen3-8B}     & $0.05$ & $0.10$ & $0.05$ & $0.29$\\
    \texttt{Qwen2.5-Coder-32B}     & $0.45$ & $\mathbf{0.53}$ & $0.32$ & $\mathbf{0.51}$\\    
    \texttt{Qwen3-32B}     & $0.35$ & $0.24$ & $0.42$ & $0.54$\\    
    \texttt{Llama3.3-70B}     & $\mathbf{0.54}$ & $0.32$& $\mathbf{0.55}$ & $0.15$\\    
    \hline
  \end{tabular}
  \caption{Model performance with RB prompt styles (FSG: Function Signature, FSC: Function Schematic), evaluated on SB (simple) and RB (regular) boards.}
  \label{tab:results_models_rb_variations}
\end{table}

\subsection{Quantitative Analysis}
\label{subsec:quantitative-analysis}

\paragraph{Performance based on Board Type}
Table~\ref{tab:results_ha} presents model accuracy on simple and regular boards for human-authored instructions. Most models perform better on simple boards. \texttt{Llama3.3-70B} and \texttt{Qwen2.5-Coder-32B} achieve higher scores of $0.69$ and $0.68$, respectively. While the scores also improve for regular boards, the gains are smaller. \texttt{Qwen2.5-Coder-32B} achieves $0.54$ accuracy. \texttt{LLaMA3.1-8B} and \texttt{Qwen3-8B} have lower scores, primarily due to hallucinations or failure to follow the expected response format, resulting in aborted executions (see abort rate in Table~\ref{tab:results_ha}). Although fine-tuning improves accuracy, the gap with the human upper bound remains substantial.

Smaller models (\texttt{Llama3.1-8B}, \texttt{Qwen3-8B}) perform better on regular boards than on simple boards. This may be due to differences in the expected output structure. For regular boards, the target code typically consists of nested \texttt{for} loops that repeat the same object placement across multiple locations (see Figure~\ref{fig:instruction_type_example_rb}). In contrast, simple boards often require placing a sequence of different shapes with precise spatial relationships, typically implemented using abstract function definitions. This output structure appears to be harder for these models to generate reliably. 

\begin{table}
  \centering
  \footnotesize
  \begin{tabular}{lcccc}
    \hline
    \multirow{2}{*}{\textbf{Model}} & \multicolumn{2}{c}{\textbf{FSG}} & \multicolumn{2}{c}{\textbf{FSC}} \\
    & SB & RB & SB & RB \\
    \hline
    \texttt{Qwen2.5-Coder-7B}     & $0.00$ & $0.00$ & $0.23$ & $0.19$\\
    \texttt{Llama3.1-8B}     & $0.01$ & $0.00$ & $0.31$ & $0.00$\\
    \texttt{Qwen3-8B}     & $0.65$ & $0.51$ & $0.55$ & $0.02$\\
    \texttt{Qwen2.5-Coder-32B}     & $0.00$ & $0.00$ & $0.00$ & $0.00$\\    
    \texttt{Qwen3-32B}     & $0.00$ & $0.00$ & $0.01$ & $0.00$\\    
    \texttt{Llama3.3-70B}     & $0.00$ & $0.00$& $0.00$ & $0.52$\\    
    \hline
  \end{tabular}
  \caption{Effect of RB prompt styles (FSG: Function Signature, FSC: Function Schematic) on abort rates for SB (simple) and RB (regular) boards.}
  \label{tab:results_haresults_models_rb_variations_abort}
\end{table}

\paragraph{Effect of prompt variation} Additional experiments were conducted using different prompt styles for regular boards. Table~\ref{tab:results_models_rb_variations} shows results from two styles: Function Signature (FSG), which includes only the function signature in the prompt, and Function Schematic (FSC), which includes the signature along with a brief schematic description of the function (see Appendix~\ref{subsec:appendix-prompt-variation} for more details). These prompt variations do not significantly affect performance on regular boards, but they reduce performance on simple boards. For example, \texttt{Qwen2.5-Coder-32B} drops from $0.68$ to $0.45$ (FSG) and $0.32$ (FSC), while \texttt{LLaMA3-70B} drops from $0.69$ to $0.54$ (FSG) and $0.55$ (FSC). These results suggest that while FSG and FSC are sufficient to support symbolic reuse (as in regular boards), they are less effective when the model is required to generate symbolic abstractions, as in simple boards.

\begin{table}
  \centering
  \small  
  \begin{tabular}{lcccc}
    \hline
    \multirow{2}{*}{\textbf{Model}} & \multicolumn{2}{c}{\textbf{SB}} & \multicolumn{2}{c}{\textbf{RB}} \\
    & ST & HA & ST & HA \\
    \hline
    \texttt{Qwen2.5-Coder-7B}     & {0.98} & {0.40} & {1.00} & {0.35}  \\
    \texttt{Llama3.1-8B}          & {0.46} & {0.08} & {0.75} & {0.21} \\
    \texttt{Qwen3-8B}            & {0.71} & {0.09} & {0.65} & {0.12} \\
    \texttt{Qwen2.5-Coder-32B}     & {0.98} & {0.68} & {1.00} & \textbf{0.54}  \\
    \texttt{Qwen3-32B}             & {0.99} & {0.53} & {1.00} & {0.38} \\
    \texttt{Llama3.3-70B}     & {0.98} & \textbf{0.68} & {1.00} & {0.33}  \\
    \hline    
    \texttt{GPT-4o}     & {0.78} & {0.72} & {0.64} & {0.55} \\
    \texttt{Claude-4}     & {0.95} & {0.88} & {0.98} & {0.60} \\    
    \hline
    \texttt{Human-Baseline}     & \multicolumn{2}{c}{{0.98}} & \multicolumn{2}{c}{\textbf{0.76}} \\ 
    % \texttt{Human-Baseline}     & {-} & {0.98} & {-} & {0.76} \\ 
    \hline    
  \end{tabular}
  \caption{Performance by instruction type: ST (synthetic) and HA (human-authored) on SB (simple) and RB (regular) boards.}
  \label{tab:results_compare_synthetic_humanwritten}
  \vspace*{-.3cm} 
\end{table}

\begin{figure*}
\centering
  \includegraphics[scale=0.65]{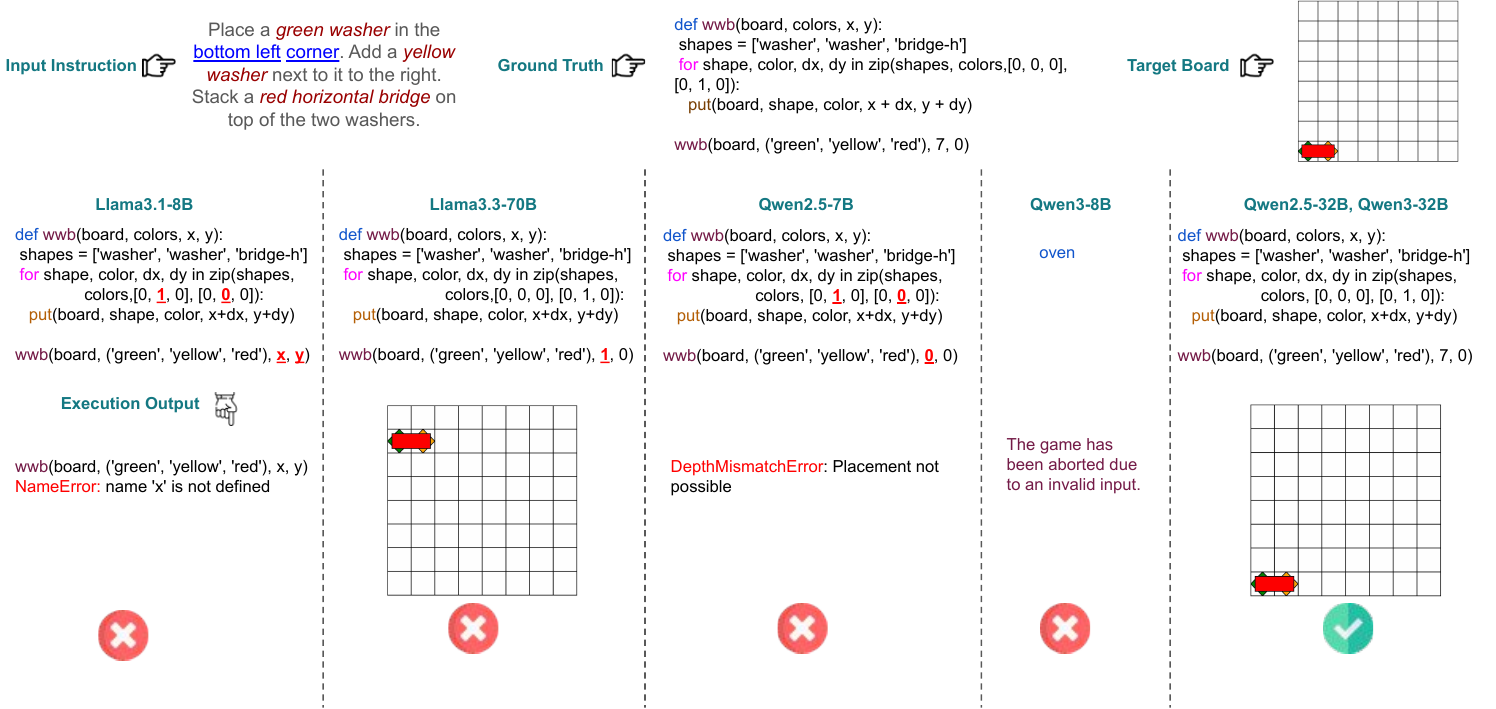}
  \caption{Execution analysis of different models on an instruction-to-code generation task. Given a natural language instruction (top left), only \texttt{Qwen2.5-32B} and \texttt{Qwen3-32B} generate correct and executable code that replicates the target board (top right). Other models fail due to issues such as undefined variables, incorrect coordinates, or invalid placements.}
  \label{fig:results_qualitatitative}
\end{figure*}

\paragraph{Few-shot prompting} We also experimented with few-shot prompting using the fine-tuned models. Table~\ref{tab:results_fewshot_rb} (in Appendix~\ref{subsubsec:appendix-fewshot-prompting}) shows that few-shot prompting improves \texttt{Qwen2.5-Coder-32B}’s score on simple boards to $0.82$, but causes a slight decrease in RB performance, from $0.54$ to $0.51$.

\paragraph{Performance based on Instruction Type} We compare model performance on synthetic and human-authored instructions to quantify the generalization gap. Table~\ref{tab:results_compare_synthetic_humanwritten} shows that while models achieve high accuracy on synthetic instructions, performance drops significantly on human-authored inputs. This discrepancy highlights the challenge of transferring symbolic reasoning skills to more abstract and natural instruction styles.

The improved regular board performance seen in \texttt{Llama3.1-8B}, \texttt{Qwen3-8B} may stem from the relative simplicity of generating repetitive structure, particularly when trained on synthetic instructions that explicitly describe iteration patterns.

Overall, fine-tuning on synthetic data improves performance on regular boards across all models. For simple boards, the impact is varied: some models, particularly instruction-tuned or code-oriented ones, benefit substantially, while others show little improvement due to format violations and hallucinations. These findings indicate that even a small amount of synthetic supervision can be beneficial, but the downstream effects depend on model alignment, size, and prompt structure.

\subsection{Qualitative Analysis}
\label{subsec:qualitative-analysis}
Figure~\ref{fig:results_qualitatitative} presents a qualitative comparison of model predictions for a human-authored instruction. The instruction requires \textit{placing a green washer in the bottom left corner of a grid, adding a yellow washer next to it on the right, and stacking a red horizontal bridge on top of the two washers}. The corresponding ground truth code places the shapes starting from (7, 0) using relative spatial offsets.

\texttt{Qwen2.5-32B} and \texttt{Qwen3-32B} generated code that exactly matched the target representation. These models correctly interpreted the spatial reference: \textit{bottom left} as (7, 0), maintained the ordering of shape placements, and generated a semantically translated abstract function.

\texttt{Llama3-70B} generated syntactically well-formed code with appropriate functional abstraction, but misinterpreted the spatial reference, mapping ``bottom left'' to (1, 0) rather than (7, 0). \texttt{Llama3-8B} shows similar behavior of correct abstraction but incorrectly uses symbolic variables (as x, y) for cell positions. \texttt{Qwen2.5-7B} generated correct structure but applied incorrect relative positions, leading to an execution-time placement error. \texttt{Qwen3-8B} generated an unrelated response, indicating failure in instruction following. This suggests that while some models are able to replicate correctly, others struggle with grounding language in spatial coordinates.

\paragraph{Error Categorization} Each model-generated code was executed, and errors were grouped based on either execution failed or succeeded with a semantic mismatch. Responses that failed to execute were classified as board placement errors and for successfully executed code, we compare the resulting board with the ground truth and identify mismatches in shape type, color, or placement order. These are classified as element mismatches.

\begin{figure}[t]
\centering
  \includegraphics[width=0.47\textwidth]{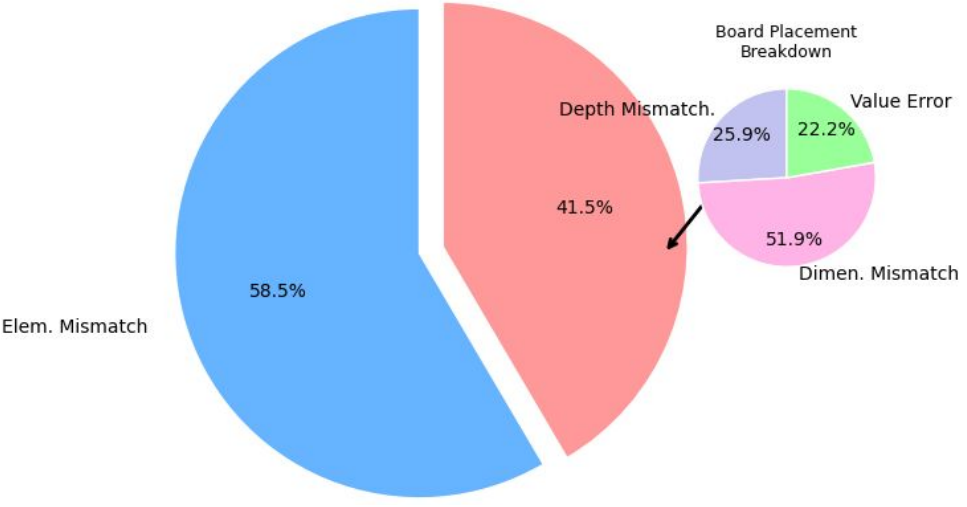}
  \caption{Error distribution on SB (simple) boards for the \texttt{Qwen2.5-Coder-32B} model after fine-tuning. The primary pie chart shows the proportion of overall error types, with ``Board Placement'' accounting for 41.5\% and 1 ``Element. Mismatch'' 58.5\%. A secondary pie further decomposes the Board Placement errors into Dimensions mismatch (51.9\%), Depth Mismatch (25.9\%), and Value Error (22.2\%)..}
  \label{fig:error_cat_sb_ft}
\end{figure}

\begin{figure}[t]
\centering
  \includegraphics[width=0.47\textwidth]{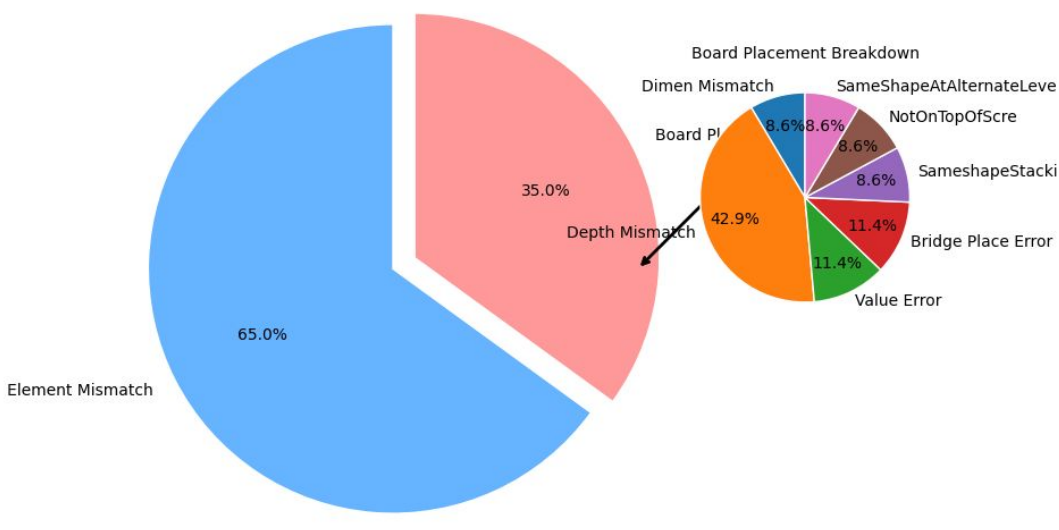}
  \caption{Error distribution on RB (regular) boards for the \texttt{Qwen2.5-Coder-32B} model after fine-tuning.}
  \label{fig:error_cat_rb_ft}
\end{figure}

Within board placement errors, we observed different error categories (see Figure~\ref{fig:error_cat_sb_ft} and Figure~\ref{fig:error_cat_rb_ft}). \textit{DepthMismatch} occurs when a bridge is placed without a supporting shape beneath it, typically due to earlier placement errors. \textit{BridgePlacement} errors involve stacking bridges above the allowed two-level height limit. \textit{DimensionMismatch} and \textit{ValueError} arise when object placements exceed board boundaries or use invalid coordinates. Other errors, such as \textit{NameError} and \textit{KeyError}, result from the use of undefined variables or incorrect dictionary access (see Figure~\ref{fig:results_qualitatitative}). Additionally, environment-specific errors such as \textit{NotOnTopOfScrew} and \textit{SameShapeStacking} reflect violations of symbolic constraints defined by the board logic.

Overall, fine-tuning reduced board placement errors across models, with most remaining errors attributable to element mismatches. (more details are in Appendix~\ref{subsec:appendix-error-categorization}).

\paragraph{Instruction Similarity} Table~\ref{tab:results_ha} reports improved model performance on human-authored instructions following fine-tuning, with larger gains observed on simple boards and smaller gains on regular boards. This difference may arise from the phrasing of regular board instructions, which often omit explicit spatial details, making it harder for models to generate the intended code. In contrast, instructions for simple boards are more concrete, even when the corresponding code is complex.

To examine the effect of instruction variation, we compute embedding similarity between synthetic and human-authored instructions for the same board configurations. We use BLEU to measure surface-level overlap and Sentence Transformers~\citep{DBLP:conf/emnlp/ReimersG19} cosine similarity to capture semantic alignment. 

Table~\ref{tab:instruction_similarity_scores} presents median and range values across board types. Instructions for simple boards show higher embedding similarity between synthetic and human-authored variants and are associated with larger execution accuracy gains. For regular boards, the similarity is lower and performance drops. To further analyze this effect, we examine how similarity varies with the number of shapes in the object. As shown in Table~\ref{tab:instruction_similarity_shapewise} (in Appendix), instructions referring to objects with fewer shapes tend to have greater execution success. These results suggest that lower semantic and lexical alignment between synthetic and human-authored instructions, particularly in structurally repetitive but abstract cases, limits the generalization ability of models fine-tuned only on synthetic data.

\begin{table}
  \centering
  \small
  \begin{tabular}{ccccc}
    \hline
    \textbf{Score Ranges} & \multicolumn{2}{c}{\textbf{BLEU}} & \multicolumn{2}{c}{\textbf{ES}} \\
    & SB & RB & SB & RB \\
    \hline
    Median Values     & {0.356} & {0.024} & {0.979} & {0.623}           \\
    Minimum Values     & {0.21} & {0.01} & {0.92} & {0.44}  \\
    Maximum Values     & {0.66} & {0.1} & {0.99} & {0.82}  \\
    \hline
  \end{tabular}
  \caption{Median and min–max values of instruction similarity (BLEU) and Embedding Similarity (ES) across Simple Boards (SB) and Regular Boards (RB). Lower ES scores on RB highlight increased linguistic variability in human instructions for complex tasks.}
  \label{tab:instruction_similarity_scores}
  \vspace*{-.5cm} 
\end{table}

\subsection{Cross-Domain Generalization}
To assess whether models generalize beyond the vocabulary and structure observed during training, we evaluate models fine-tuned on synthetic data on other datasets.

\paragraph{Transfer to Hexagons} The \textsc{HEXAGONS} dataset~\citep{lachmy-etal-2022-draw} consists of human-authored instructions for coloring specific cells on a hexagonal grid, based on natural language descriptions. These instructions are similar to repetitive patterns, resembling the regular board structures in our setup. The coloring operation, denoted as \texttt{paint(color, row, column)}, is analogous to our \texttt{put(shape, color, row, column)} function. The dataset uses a $10\times18$ hexagonal grid. For evaluation, we use the test split\footnote{\url{https://github.com/OnlpLab/Hexagons/blob/main/data/test.jsonl}}, which includes $62$ drawing procedures. Table~\ref{tab:results_dmf_tidybot} reports model performance on this task before and after fine-tuning. The fine-tuned models show no improvement compared to pretrained models. This limited transferability may be attributed to the domain shift between the training data and the \textsc{HEXAGONS} instructions, which differ in both linguistic style and task structure. Additionally, the larger grid size ($10\times18$) significantly increases the complexity of spatial reasoning and the difficulty of learning accurate coordinate mappings. 

\begin{figure}[t]
\centering
  \includegraphics[width=\columnwidth]{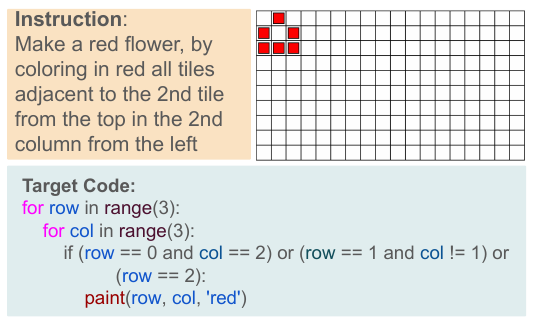}
  \caption{Example from the drawing hexagons task with a human-authored instruction and its corresponding target code.}
  \label{fig:dmf_example}
\end{figure}

\paragraph{Transfer to TidyBot} The \textsc{TidyBot} dataset~\citep{DBLP:conf/iros/WuAKLZSBRF23} contains free-form human instructions for real-world object arrangement. The core operation, expressed as \texttt{pick\_and\_place(item, newposition)}, is similar to shape placement in our setting through \texttt{put(shape, color, row, column)}. Results are presented in Table~\ref{tab:results_dmf_tidybot}. Top-performing models such as \texttt{Qwen2.5-32B} and \texttt{LLaMA3-70B} maintain high task success, demonstrating generalization to other domain instructions and object categories. This suggests that fine-tuning on structured synthetic spatial tasks supports transfer to some domains. In contrast, models such as \texttt{Llama3.1-8B}, \texttt{Qwen3-8B} show significant performance degradation, indicating limited generalization and potential overfitting to the training domain.

\begin{figure}[t]
\centering
  \includegraphics[width=\columnwidth]{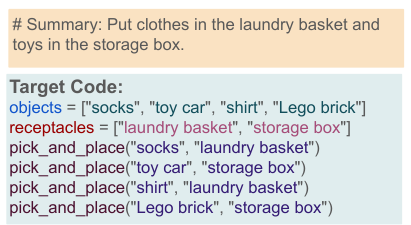}
  \caption{Example task from Tidybot with human-authored instruction and corresponding target code.}
  \label{fig:tidybot_example}
\end{figure}

\begin{table}
  \centering
  \footnotesize
  \begin{tabular}{lcccc}
    \hline
    \multirow{2}{*}{\textbf{Model}} & \multicolumn{2}{c}{\textbf{\textsc{HEXAGONS}}} & \multicolumn{2}{c}{\textbf{\textsc{TIDYBOT}}} \\
    & {Before}& {After}& {Before} & {After} \\
    \hline
    \texttt{Qwen2.5-Coder-7B}     & $0.00$ & $0.03$ & $0.00$ & $0.00$\\
    \texttt{Llama3.1-8B}     & $0.00$ & $0.00$ & $0.00$ & $0.039$\\
    \texttt{Qwen3-8B}     & $0.00$ & $0.00$ & $0.00$ & $0.00$\\
    \texttt{Qwen2.5-Coder-32B}    & $\mathbf{0.10}$ & $\mathbf{0.11}$& $\mathbf{0.82}$ & $\mathbf{0.90}$\\
    \texttt{Qwen3-32B}     & $0.00$ & $0.03$& $0.00$ & $0.00$\\
    \texttt{Llama3.3-70B}    & $0.06$ & $\mathbf{0.10}$  & $0.00$ & $0.06$\\
    \hline    
    \texttt{Human-Baseline}     & \multicolumn{2}{c}{0.76} & \multicolumn{2}{c}{0.95}\\        
    \hline    
  \end{tabular}
  \caption{Model performance on \textsc{HEXAGONS} and \textsc{TIDYBOT} tasks before and after fine-tuning.}
  \label{tab:results_dmf_tidybot}
  \vspace*{-.3cm} 
\end{table}

\section{Conclusion}
This paper investigates when and how synthetic instruction data can support generalization to human-authored instructions in grounded spatial reasoning tasks. Our findings show that synthetic-only fine-tuning can enable generalization when the arrangement does not involve object repetitions. Through semantic similarity analysis and error categorization, we identified referential ambiguity as a bottleneck: performance declines when instructions contain linguistic variations due to object repetitions. We also observe that regular boards execution success rates see no improvement even with prompt-style variations and few-shot prompting. This suggests that LLMs trained on direct, explicit instruction sets are insufficiently equipped to handle the nuances of compositional abstraction inherent in human-authored instructions. These instructions consists implicit repetition, or structural symmetry posing challenges for models that lack inductive bias toward programmatic generalization. Future work should focus on developing synthetic datasets that mimic these linguistic variations.

\section*{Limitations}

Our study focuses on generalization from synthetic to human instructions in spatial reasoning tasks, but several limitations remain. First, the training data is entirely synthetic and rule-based; while it enables controlled supervision, it lacks the linguistic diversity, ambiguity, and noise characteristic of real-world language. Second, the target code representation is highly task-specific, designed for grid-based pick-and-place operations, which may limit transferability to other spatial reasoning domains or instruction-following tasks with different semantics. Third, although our evaluation includes human-authored instructions, they are limited to single-turn settings and do not capture the challenges of multi-turn or collaborative spatial interactions.

\bibliography{acl_latex}

\appendix
\section{Appendix}
\label{sec:appendix}

\subsection {Prompt Templates}
\label{sec:appendix-prompt-templates}
In our task setup, LLMs are prompted to translate natural language instructions into structured code representations. We adopt a chat-style format to align with the pretraining of instruction-tuned models and design a multi-part prompt (see Figure~\ref{fig:promptstructure_sb} and Figure~\ref{fig:promptstructure_rb}) that includes details of the $2.5$D grid environment along with available API functions. The prompt structure differs slightly between \textit{Simple} and \textit{Regular} boards—particularly in terms of response constraints and the nature of in-context examples provided. This same prompt format is used both during inference and fine-tuning.

\begin{figure}
\centering
  \includegraphics[width=0.48\textwidth]{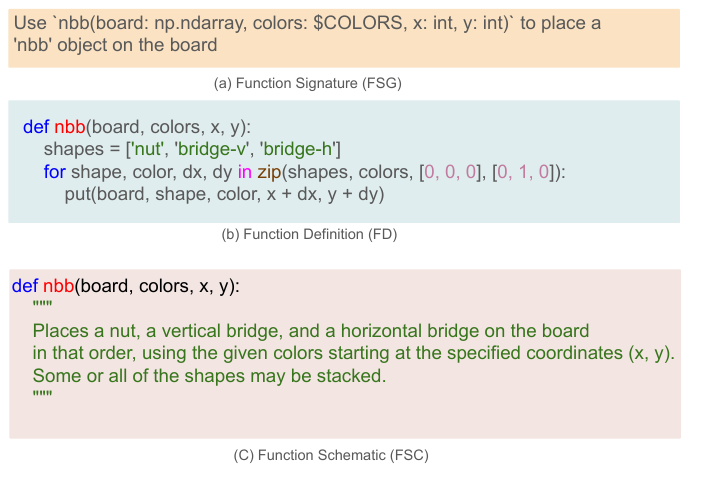}
  \caption{Different styles of prompt variations used for regular board instructions during training and inference.}
  \label{fig:prompt_variations_funcdef}
\end{figure}

\subsection{Prompt Variations}
\label{subsec:appendix-prompt-variation}

To explore how the structure of model prompts affects performance, we introduce three prompt variations for regular board instructions. These variations differ in how the abstract function used in the instruction (e.g., \textit{nbb}) is presented in the prompt context. All three settings include the instruction and the expected output. The difference lies in the prompt information provided about the nbb function. These prompt settings were used consistently across both training and evaluation for each variation. Their impact on model performance is discussed in Section~\ref{subsec:quantitative-analysis}, and associated results are reported in Table~\ref{tab:results_models_rb_variations}.
\paragraph{Function Signature (FSG):} The prompt includes only the function signature. This setting simulates the case where the model is expected to use a known abstract function without being shown its implementation. The intent is to assess whether the model can learn to infer the correct usage pattern based solely on its name.

\paragraph{Function Definition (FD):}
The full definition of the function is included in the prompt. This setting provides the most information about what the function does. It is intended to reduce ambiguity during training and inference and help models correctly ground object placement logic based on a concrete implementation.

\begin{table}
  \centering
  \small
  \begin{tabular}{lcccc}
    \hline
    \multirow{2}{*}{\textbf{Model}} &\multicolumn{2}{c}{\textbf{ST}} & \multicolumn{2}{c}{\textbf{HA}} \\
    & {\textbf{SB}} & {\textbf{RB}} & {\textbf{SB}} & {\textbf{RB}} \\
    \hline
    \texttt{Qwen2.5-Coder-7B}     & $0.95$ & $0.73$ & $0.45$ & $0.20$ \\
    \texttt{Llama3.1-8B}     & $0.00$ & $0.00$ & $0.00$ & $0.02$ \\
    \texttt{Qwen3-8B}          & $0.03$ & $0.29$ & $0.01$ & $0.10$ \\
    \texttt{Qwen2.5-Coder-32B}  & $0.97$ & $0.98$  & $\mathbf{0.82}$ & $\mathbf{0.51}$ \\    
    \texttt{Qwen3-32B}          & $0.89$ & $0.52$ & $0.49$ & $0.27$ \\
    \texttt{Llama3.3-70B}     & $0.73$ & $0.00$ & $0.30$& $0.07$  \\    
    \hline
  \end{tabular}
  \caption{Task performance (accuracy) on simple boards (SB) and regular boards (RB) for synthetic (ST) instructions and human-authored (HA) instructions using few-shot prompting with the Fine-tuned models. Except for \texttt{Qwen2.5-Coder-32B} all the models performance was detoriated.}
  \label{tab:results_fewshot_rb}
\end{table}

\paragraph{Function Schematic (FSC):}
The prompt includes the function signature and a schematic docstring describing the function’s purpose, but not the actual implementation. This setup serves as an intermediate between FSG and FSD. The schematic is intended to provide semantic guidance without dictating specific implementation. It tests whether models benefit from lightweight documentation, which mirrors how humans often interact with partially known APIs.

\subsubsection{Few-shot Prompting}
\label{subsubsec:appendix-fewshot-prompting}
To evaluate whether performance can be further improved with in-context examples, we experimented with few-shot prompting at inference time using the fine-tuned models. For each evaluation instance, we added five synthetic instruction–output pairs of the same board type (simple or regular) to the prompt, followed by the test instruction. The examples were randomly sampled from the training data but excluded any exact match with the test input.

Table~\ref{tab:results_fewshot_rb} reports the results. We observed that few-shot prompting improved performance only for \texttt{Qwen2.5-Coder-32B} model and its accuracy is improved to $0.82$ (compared to $0.68$ in zero-shot). However, the same setting led to a slight performance drop on regular boards, possibly due to increased sensitivity to syntactic variation in human-authored instructions.

\subsection{Fine-Tuning}
Our dataset consists of spatial instruction-target code pairs over two environment types: \textit{Simple} Boards (SB) and \textit{Regular} Boards (RB). Table~\ref{tab:datasetstats} summarizes the distribution across training, validation, and test splits for each board type. These splits form the basis for all fine-tuning and evaluation setups described below. All models are fine-tuned using the Unsloth~\footnote{\url{https://unsloth.ai/}} framework with a 4-bit quantization setup to reduce memory overhead. We apply parameter-efficient fine-tuning using LoRA with the following configuration: rank $r=16$, $\alpha=16$, dropout=$0.0$, batch size = $8$, learning rate = $1\mathrm{e}{-4}$, and the \texttt{adam-8bit} optimizer. This configuration was selected based on preliminary ablation experiments that balanced good trade-off between performance and training efficiency. Fine-tuning is performed for three epochs, and \textit{four} gradient accumulation steps, across all experiments. We use the same prompt template format described in Appendix~\ref{sec:appendix-prompt-templates}, ensuring consistency between training and evaluation. All experiments were conducted on a single NVIDIA A100 80GB GPU.

\begin{table} [t]
  \centering
  \begin{tabular}{lc}
    \hline
    \textbf{Dataset} & \textbf{SB} \\
    \hline
    \verb|Training Set|     & {1072}           \\
    \verb|Validation Set|     & {130}           \\
    \verb|Test Set|     & {130}           \\\hline
  \end{tabular}
  \begin{tabular}{lc}
    \hline
    \textbf{RB} \\
    \hline
    {1168}          \\
    {130}          \\
    {130}            \\
    \hline
  \end{tabular}
  \begin{tabular}{lc}
    \hline
    \textbf{TB} \\
    \hline
    {96}           \\
    {-}           \\
    {96}           \\\hline
  \end{tabular} 
  \caption{Dataset statistics showing the number of boards in Simple Boards (SB), Regular Boards (RB), and number of scenarios in TidyBot (TB) across training, validation, and test splits.}
  \label{tab:datasetstats}
\end{table}

\subsubsection{Ablation Study}
\label{subsubsec:appendix-finetuning-ablationstudy}
We conducted multiple ablation studies focused on finding the suitable hyper parameter configuration. We begin with identifying the impact of learning rate and batch size, keeping the number of training epochs fixed at 3. We compared learning rates of $1\mathrm{e}{-4}$ and $2\mathrm{e}{-4}$, across batch sizes of 2, 4 and 8. We observed that higher learning rates ($2\mathrm{e}{-4}$) led to faster convergence but sometimes caused overfitting in larger models, particularly when paired with smaller batch sizes. Validation loss curves remained stable for smaller models, but the downstream accuracy did not consistently improve. 

To further examine the effect of training duration, we varied the number of epochs while using early stopping (patience 2). We compared runs with 1 epoch, 2 epochs, and 3 epochs (with and without early stopping). Reducing to 1 epoch with a higher learning rate ($2\mathrm{e}{-4}$) led to unstable gains and greater variance across models. Overall, 3 epochs with early stopping of 2 provided more robust results across both board types. 

Finally, we evaluated how many training examples were needed to achieve meaningful generalization. We fixed the model and training setup, and varied the number of fine-tuning samples (e.g., 100, 300, 500, and 1000). We observed that most models began to generalize reliably to human-authored instructions only when trained on at least 1000 examples (balanced across simple and regular boards). With fewer than 500 examples, performance degraded sharply, especially for regular boards.

\begin{table*}
  \centering
  \begin{tabular}{ccccccc}
    \hline
    \multirow{2}{*}{\textbf{Shapes Per Object}} & \multicolumn{3}{c}{\textbf{SB}} & \multicolumn{3}{c}{\textbf{RB}} \\
    & BLEU & ES & SR & BLEU & ES & SR \\
    \hline
    \texttt{2}     & $0.553$ & $0.979$ & $\mathbf{1.00}$ & $0.033$  & $0.545$ & $\mathbf{0.80}$\\
    \texttt{3}     & $0.422$ & $0.977$ & $0.90$ & $0.027$  & $0.626$ & $0.43$\\
    \texttt{4}     & $0.327$ & $0.979$ & $0.74$ & $0.024$  & $0.635$ & $0.46$\\
    \texttt{5}     & $0.275$ & $0.979$ & $0.29$ & $0.021$  & $0.671$ & $0.59$\\    
    \hline
   \end{tabular}
  \caption{Instruction similarity and execution success rate (SR) for \texttt{Qwen2.5-Coder-32B}, grouped by number of shapes per object. Higher shape counts correlate with reduced lexical and semantic similarity between human and synthetic instructions, and lower execution success.; ES: Embedding Similarity Score;}
  \label{tab:instruction_similarity_shapewise}
\end{table*}    

\subsection {Instruction Similarity vs. Execution Success}
\label{sec:appendix-semantic-scores}
We compute similarity between synthetic and human instructions associated with a given board configurations using two measures: BLEU (surface-level overlap) and Embedding cosine similarity using Sentence-BERT (semantic-level alignment).

Table~\ref{tab:instruction_similarity_shapewise} shows how the semantic and lexical similarity between synthetic and human-authored instructions varies with the number of shapes involved in the described object. We observe a clear trend: as the number of shapes per object increases, both lexical and semantic similarity between synthetic and human-written instructions decreases. This decrease is mirrored by a drop in execution success rate (SR).

For simple boards, even though the instructions are often more explicit than those for regular boards, the primary challenge lies in mapping the linguistic description to precise spatial relations and composing an appropriate abstract function. When more shapes are involved, spatial dependencies become more complex, and function construction becomes harder. This leads to a significant decline in execution success, from $1.00$ for two-shape objects to $0.29$ for five-shape objects.

For regular boards, the instructions often describe repetitive object placement patterns using abstract or minimal phrasing (e.g., ``repeat in the next row''). As the number of shapes in the object increases, the underlying pattern logic also becomes more complex, which complicates code generation, particularly when the model must correctly interpret spatial regularities not explicitly stated in the instruction. While the SR remains higher than for simple boards in some cases, the same downward trend is observed.

These findings suggest that both the semantic gap between synthetic and human instructions and the object complexity (in terms of shape count and spatial arrangement) jointly influence model generalization. Models fine-tuned only on synthetic data show limited ability to bridge this gap as object complexity increases.

\begin{comment}
\begin{table*}
  \centering
  \begin{tabular}{lcccccccc}
    \hline   
    \multirow{3}{*}{\textbf{Model}} &  \multicolumn{4}{c}{\textbf{SB}} & \multicolumn{4}{c}{\textbf{RB}} \\
    & \multicolumn{2}{c}{\textbf{EM}} & \multicolumn{2}{c}{\textbf{BP}} & \multicolumn{2}{c}{\textbf{EM}} & \multicolumn{2}{c}{\textbf{BP}}\\
    & \textbf{Before} & \textbf{After} & \textbf{Before} & \textbf{After} & \textbf{Before} & \textbf{After} & \textbf{Before} & \textbf{After}\\
    \hline    
    \texttt{Qwen2.5-Coder-7B}     & $50.70$ & $50.00$ & $49.30$ & $50.00$ & $49.30$ & $0.05$ & $0.00$ & $1.00$\\

    \texttt{Llama3.1-8B}     & $2.40$ & $38.5$ & $95.30$ & $49.50$ & $97.70$ & $0.05$ & $0.00$ & $1.00$\\

    \texttt{Qwen3-8B}     & $40.00$ & $49.30$ & $60.00$ & $50.70$ & $60.00$ & $0.05$ & $0.00$ & $1.00$\\

    \texttt{Qwen2.5-Coder-32B}   & $54.20$ & $35.70$ & $45.80$ & $64.30$ & $45.80$ & $0.05$ & $0.00$ & $1.00$\\

    \texttt{Qwen3-32B}     & $42.00$ & $49.20$ & $54.00$ & $50.80$ & $58.00$ & $0.05$ & $0.00$ & $1.00$ \\    
    \texttt{Llama3.3-70B}     & $7.40$ & $60.00$ & $92.60$ & $40.00$ & $64.30$ & $83.50$ & $35.70$ & $16.50$\\   
    \hline
  \end{tabular}
  \caption{Error categorization before and after fine-tuning on SB (simple) and RB (regular) boards. \todo{value update}}
  \label{tab:results_error_categorization}
\end{table*}
\end{comment}

\begin{table*}
  \centering
  \footnotesize
  \begin{tabular}{lcccc|cccc}
    \hline
    \multirow{3}{*}{\textbf{Model}} & \multicolumn{4}{c}{\textbf{Board Placement $\downarrow$}} & \multicolumn{4}{c}{\textbf{Element Mismatch $\downarrow$}} \\
    & \multicolumn{2}{c}{\textbf{SB}} & \multicolumn{2}{c}{\textbf{RB}} & \multicolumn{2}{c}{\textbf{SB}} & \multicolumn{2}{c}{\textbf{RB}}
    \\
    & Before & After  & Before & After & Before & After & Before & After \\
    \hline
    \texttt{Qwen2.5-Coder-7B}     & $80.17$ & $50.00$ & $-$ & $24.00$ & $19.83$ & $50.00$ & $-$ & $76.47$ \\
    \texttt{Llama3.1-8B}     & $55.00$ & $61.56$ & $45.00$ & $28.41$ & $45.00$ & $38.46$ & $0.00$ & $71.59$ \\
    \texttt{Qwen3-8B}          & $-$ & $50.72$ & $100.00$ & $66.67$ & $-$ & $49.28$ & $-$ & $33.33$\\
    \texttt{Qwen2.5-Coder-32B}  & $44.64$ & $64.29$  & $33.65$ & $20.00$ & $55.36$ & $35.71$ & $66.35$ & $80.00$\\    
    \texttt{Qwen3-32B}          & $100.00$ & $50.82$ & $100.00$ & $25.93$ & $-$ & $49.18$ & $-$ & $74.07$\\
    \texttt{Llama3.3-70B}     & $78.45$ & $40.00$ & $14.29$& $16.46$ & $21.56$ & $60.00$ & $85.71$ & $83.54$ \\      
    \hline    
  \end{tabular}
  \caption{Error Categorization across nodel responses before and after fine-tuning}
  \label{tab:results_ha_error_categorization}
\end{table*}
\input
\subsection {Error Categorization}
\label{subsec:appendix-error-categorization}
Table~\ref{tab:results_ha_error_categorization} presents the percentage of error types observed in model outputs, both before and after fine-tuning. For some models, entries are marked with a dash ('-'), indicating a 100\% abort rate and the absence of usable responses for error categorization.

The percentages are computed based on the total number of errors extracted from the model outputs, categorized into Board Placement and Element Mismatch errors.

Before fine-tuning, most models exhibit a higher proportion of Board Placement errors. This is often due to spurious generations violating environmental constraints, such as exceeding spatial boundaries or ignoring depth ordering. After fine-tuning, however, Element Mismatch becomes the dominant error type—suggesting that while structural placement improves, models still struggle to accurately map the intended elements to their correct spatial positions.

The table also helps to qualitatively assess the overall trend in error composition and model behavior across fine-tuning stages.

\begin{figure*}
\centering
  \includegraphics[scale=0.68]{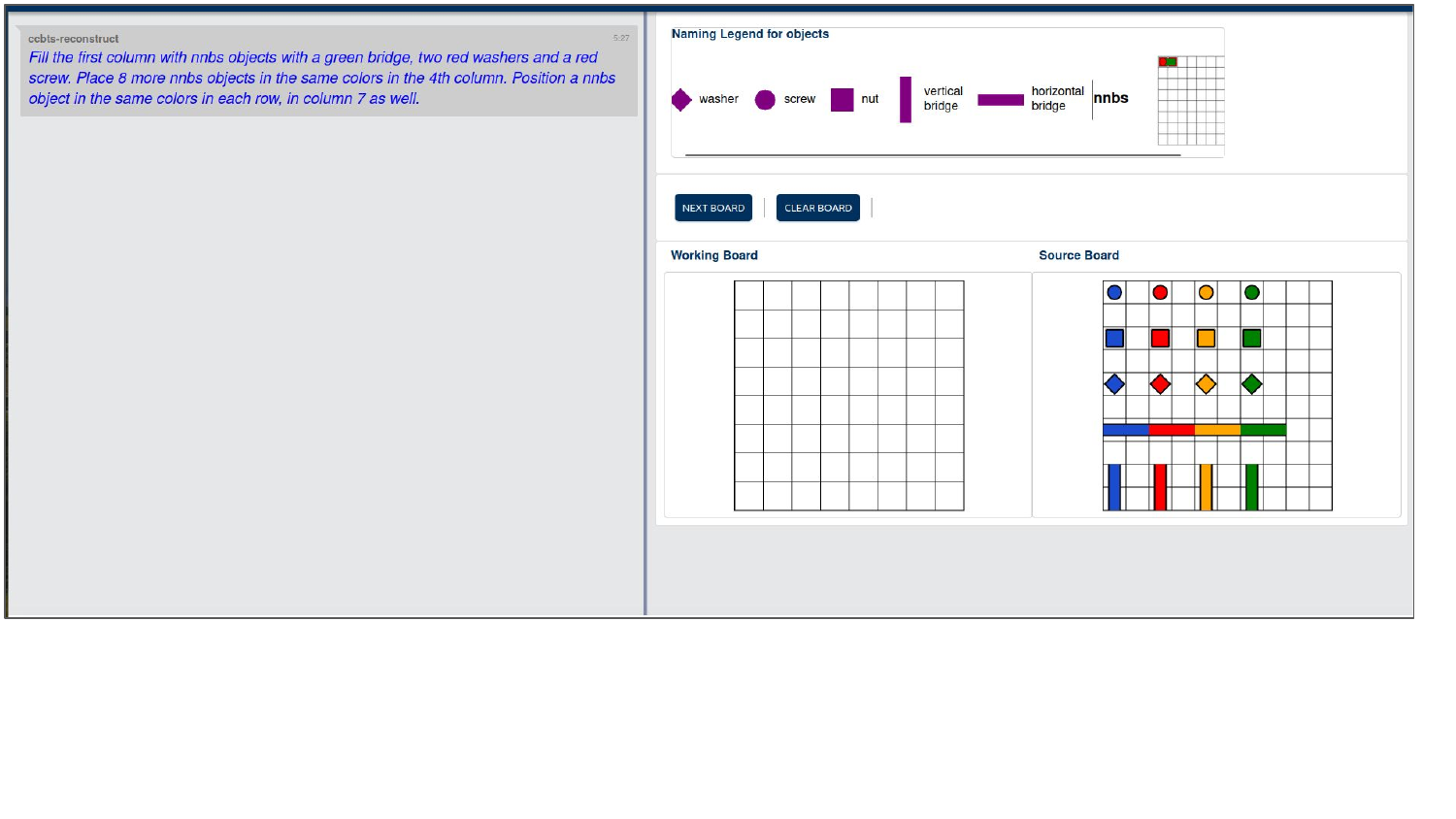}
  \caption{Web interface used for human reconstruction of spatial configurations. Instructions appear on the left, while the right panel allows annotators to place objects on an 8×8 grid using a visual palette. The interface supports actions such as object selection and cell duplication to aid in building complex or repetitive structures.}
  \label{fig:human_reconstruct_interface}
\end{figure*}

\subsection {Human Evaluation}
\label{sec:appendix-humaneval}
To establish a human performance baseline, we developed an interactive web interface (Figure~\ref{fig:human_reconstruct_interface}) that displays the instruction on the left and an editable 8$\times$8 grid on the right. Annotators were tasked with reconstructing the spatial configuration described by the input instruction using the grid. The interface included features such as cell copy-paste to efficiently handle repeated structures and embedded step-by-step guidelines to standardize the reconstruction process.

An annotator was recruited to perform the task. While the annotator had no prior experience in spatial layout tasks, they were familiar with general data labeling workflows. The evaluation involved reconstructing a total of 260 boards (130 \textit{Simple} and 130 \textit{Regular}) and required approximately 30 hours to complete.

The reconstructed boards were programmatically (by executing the constructed scripts as a result of interaction) compared to the corresponding gold configurations. The resulting match scores are reported as human baseline accuracies and serve as an upper bound reference for model performance.

\begin{figure*}
  \centering
  %\begin{subfigure}{\textwidth}
  %  \centering  
    \begin{prompt}
\\
\textbf{System Info}
\\\\
You are a helpful assistant who is designed to interpret and translate natural language instructions into python executable code snippets.
\\\\
\textbf{Environment Info}
\\\\
The environment is an 8x8 grid allowing shape placement and stacking. A shape can be placed in any cell, while stacking involves adding multiple shapes to the same cell, increasing its depth. Shapes typically occupy a single cell, except for the "bridge," which spans two cells and requires two other shapes for stacking. Horizontal bridges span adjacent columns (left and right), and vertical ones span consecutive rows (top and bottom). Stacking is only possible if the shapes have matching depths.
\\\\
In the grid, columns align with the x-axis and rows with the y-axis. Python indexing is used to identify each cell. The cell in the top-left corner is in the first row and first column, corresponding to x and y values of 0, 0. Similarly, the top-right corner cell is in the first row and eighth column, with x and y values of 0, 7.
\\
- Use the shape name 'bridge-h' if a bridge is placed horizontally
\\
- Use the shape name 'bridge-v' if a bridge is placed vertically
\\\\
The following functions are already defined; therefore, do not generate additional code for it
\\
- Use `put(board: np.ndarray, shape: string, color: string, x: int, y: int) to place a shape on the board
\\\\
\textbf{Task Info}
\\\\
For each instruction labeled Instruction: please respond with code under the label Function: followed by a newline and usage for the function under the label Usage: followed by a newline.
\\\\
\textbf{Context Info}
\\\\
\$INCONTEXT\_SAMPLES
\\\\
\textbf{Other Info}
\\\\
Do not generate any other text/explanations.
\\\\
The order of colors, x, y matters, as these are assigned to the shapes in the same sequence.
Ensure the response can be executed by Python `exec()`, e.g.: no trailing commas, no periods, etc.
\\
Let's begin
\\\\
Instruction:
\\
\$TEST\_INSTRUCTION
\end{prompt}
\caption{Prompt template used for the spatial grounding task for \textit{Simple} boards. The system information specifies system level behavior, the environment information indicates the environment details of the user-agent environment, the context information describes the in-context examples, task information indicates the specific response format to follow.}
    \label{fig:promptstructure_sb}
  %\end{subfigure}
\end{figure*}
\begin{figure*}
  \centering
  %\begin{subfigure}{\textwidth}
  %  \centering  
    \begin{prompt}
\\
\textbf{System Info}
\\\\
You are a helpful assistant who is designed to interpret and translate natural language instructions into python executable code snippets.
\\\\
\textbf{Environment Info}
\\\\
The environment is an 8x8 grid allowing shape placement and stacking. A shape can be placed in any cell, while stacking involves adding multiple shapes to the same cell, increasing its depth. Shapes typically occupy a single cell, except for the "bridge," which spans two cells and requires two other shapes for stacking. Horizontal bridges span adjacent columns (left and right), and vertical ones span consecutive rows (top and bottom). Stacking is only possible if the shapes have matching depths.
\\\\
In the grid, columns align with the x-axis and rows with the y-axis. Python indexing is used to identify each cell. The cell in the top-left corner is in the first row and first column, corresponding to x and y values of 0, 0. Similarly, the top-right corner cell is in the first row and eighth column, with x and y values of 0, 7.
\\\\
- Use the shape name 'bridge-h' if a bridge is placed horizontally
- Use the shape name 'bridge-v' if a bridge is placed vertically
\\\\
The following functions are already defined; therefore, do not generate additional code for it
\\
- Use `put(board: np.ndarray, shape: string, color: string, x: int, y: int)` to place a shape on the board
\\
- Use `\$COMBO\_NAME(board: np.ndarray, colors: \$COLORS, x: int, y: int)` to place a '\$COMBO\_NAME' object on the board
\\\\
\textbf{Task Info}
\\\\
For each instruction labeled Instruction: please respond with code under the label Output: followed by a newline.
\\\\
\textbf{Context Info}
\\\\
\$INCONTEXT\_SAMPLES
\\\\
\textbf{Other Info}
\\\\
Do not generate any other text/explanations.
\\\\
The order of colors, x, y matters, as these are assigned to the shapes in the same sequence.
Ensure the response can be executed by Python `exec()`, e.g.: no trailing commas, no periods, etc.
\\
Lets begin
\\\\
Instruction:
\\
\$TEST\_INSTRUCTION
\end{prompt}
\caption{Prompt template used for the spatial grounding task for \textit{Regular} boards. The system information specifies system level behavior, the environment information indicates the environment details of the user-agent environment, the context information describes the in-context examples, task information indicates the specific response format to follow.}
    \label{fig:promptstructure_rb}
  %\end{subfigure}
\end{figure*}

\end{document}